\documentclass[journal, a4paper,12pt,onecolumn]{IEEEtran}
\usepackage{graphicx}
\usepackage[english]{babel}
\usepackage[latin1]{inputenc}
\usepackage{amssymb,amsmath}
\usepackage{url}
\usepackage{cite}

\usepackage{lineno}

\usepackage{setspace}

\usepackage{multicol}
\usepackage{color,comment}
 \usepackage{chngpage}
 \usepackage{url}

\begin{document}
%\linenumbers

\title{Adaptive imputation of missing values for incomplete pattern classification}
\author{{Zhun-ga Liu$^{1}$, Quan Pan$^1$, Jean Dezert$^2$, Arnaud Martin$^3$}
  
{ 1. School of Automation,
 Northwestern Polytechnical University, Xi'an, China.\\   
Email: liuzhunga@nwpu.edu.cn\\
 2. ONERA - The French Aerospace Lab, F-91761 Palaiseau, France.\\
 Email: jean.dezert@onera.fr\\
 3.IRISA, University of Rennes 1, Rue E. Branly, 22300 Lannion, France.\\
 Email: Arnaud.Martin@univ-rennes1.fr
} }

\maketitle

\vspace{-1cm}
\begin{abstract}
In classification of incomplete pattern, the missing values can either play a crucial role in the class determination, or have only little  influence (or eventually none) on the classification results according to the context. We propose a credal classification method for incomplete pattern with adaptive imputation of missing values based on belief function theory. At first, we try to classify the object (incomplete pattern) based only on the available attribute values. As underlying principle, we assume that the missing information is not crucial for the classification if a specific class for the object can be found using only the available information. In this case, the object is committed to this particular class. However, if the object cannot be classified without ambiguity, it means that the missing values play a main role for achieving an accurate classification. In this case, the missing values will be imputed based on the $K$-nearest neighbor (K-NN) and self-organizing map (SOM) techniques, and the edited pattern with the imputation is then classified. The (original or edited) pattern is respectively classified according to each training class, and the classification results represented by basic belief assignments are fused with proper combination rules for making the credal classification. The object is allowed to belong with different masses of belief to the specific classes and meta-classes (which are particular disjunctions of several single classes).  The credal classification captures well the uncertainty and imprecision of classification, and reduces effectively the rate of misclassifications thanks to the introduction of meta-classes. The effectiveness of the proposed method with respect to other classical methods is demonstrated based on several experiments using artificial and real data sets.
\end{abstract}


%\IEEEpeerreviewmaketitle
%===================================
\section{Introduction}
%%%=================================

In many practical classification problems, the available information for making object classification is partial (incomplete) because some attribute values can be missing due to various reasons (e.g. the failure or dysfunctioning of the sensors providing information, or partial observation of object of interest because of some occultation phenomenon, etc). So it is crucial to develop efficient techniques to classify as best as possible the objects with missing attribute values (incomplete pattern), and the search for a solution of this problem remains an important research topic in the pattern classification field \cite{review, random}. Some more details about pattern classification can be found in \cite{PR, PRML}.

There have been many approaches developed for classifying the incomplete patterns \cite{review}, and they can be broadly grouped into four different types.
The first (simplest) one is to remove directly the patterns with missing values, and the classifier is designed only for the complete patterns. This method is acceptable when the incomplete data set is only a very small subset (e.g. less than 5\%) of the whole data set, but it cannot effectively classify the pattern with missing values.
 The second type is the model-based techniques \cite{MoI}. The probability density function (PDF) of the input data (complete and incomplete cases) is estimated at first by means of some procedures, and then the object is classified using bayesian reasoning. For instance, the expectation-maximization (EM) algorithm have been applied to many problems involving missing data for training Gaussian mixture models  \cite{MoI}. In the model-based methods, it must make assumptions about the joint distribution of all the variables in the model, but the suitable distributions sometimes are hard to obtain.
The third type classifiers are designed to directly handle incomplete pattern without imputing the missing values, such as neural network ensemble methods \cite{NNEM}, decision trees \cite{DT}, fuzzy approaches \cite{FCMCI} and support vector machine classifier \cite{MSVM}.
 The last type is the often used imputation (estimation) method. The missing values are filled with proper estimations \cite{ImI} at first, and then the edited patterns are classified using the normal classifier (for the complete pattern). The missing values and pattern classification are treated separately in these methods. Many works have been devoted to the imputation of missing data, and the imputation can be done either by the statistical methods, e.g. mean imputation \cite{MI}, regress imputation \cite{random}, etc, or by machine learning methods, e.g. K-nearest neighbors imputation (KNNI) \cite{knnI}, Fuzzy $c$-means (FCM) imputation (FCMI) \cite{FCMI1, FCMI2}, Self-organizing map imputation (SOMI) \cite{SOMI}, etc. In KNNI, the missing values are estimated using K-nearest neighbors of object in training data space. In FCMI, the missing values are imputed according to the clustering centers of FCM and taking into account the distances of the object to these centers \cite{FCMI2, FCMI1}. In SOMI \cite{SOMI}, the best match node (unit) of incomplete pattern can be found ignoring the missing values, and the imputation of the missing values is computed based on the weights of the activation group of nodes including the best match node and its close neighbors.
These existing methods usually attempt to classify the object into a particular class with maximal probability or likelihood measure. However, the estimation of missing values is in general quite uncertain, and the different imputations of missing values can yield very different classification results, which prevent us to correctly commit the object into a particular class.

Belief function theory (BFT), also called Dempster-Shafer theory (DST) \cite{Shafer} and its extension \cite{ TBM1,Jean} offer a mathematical framework for modeling uncertainty and imprecise information \cite{Jousselme2006}. BFT has already been applied successfully for object classification \cite{eknn,knnoptimi, case-model, BKNN, martin1, ANNbf, CCR, PCC, Deng1}, clustering \cite{BCM,CCM, MECM, ECM,em} and multi-source information fusion \cite{weightedcombination, DSD, Li1, Han1}, etc. Some classifiers for the complete pattern based on DST have been developed by Den{\oe}ux and his collaborators to come up with the evidential K-nearest neighbors (EK-NN) \cite{eknn}, evidential neural network (ENN) \cite{ANNbf}, etc. The extra ignorance element represented by the disjunction of all the elements in the whole frame of discernment is introduced in these classifiers to capture the totally ignorant information. However, the partial imprecision, which is very important in the classification, is not well characterized. We have proposed credal classifiers \cite{BKNN,CCR} for complete pattern considering all the possible meta-classes (i.e. the particular disjunctions of several singleton classes) to model the partial imprecise information. The credal classification allows the objects to belong (with different masses of belief) not only to the singleton classes, but also to any set of classes corresponding to the meta-classes. In \cite{BKNN}, a belief-based $K$-nearest neighbor classifier (BK-NN) has been presented, and the credal classification of object is done according to the distances between the object and its $K$ nearest neighbors as well as two given (acceptance and rejection) distance thresholds. The K-NN classifier generally takes big computation burden, and this is not convenient for real application. Thus, a simple credal classification rule (CCR) \cite{CCR} has been further developed, and the belief value of object associated with different classes (i.e. singleton classes and selected meta-classes) is directly calculated by the distance to the center of corresponding class and the distinguishability degree (w.r.t. object) of the singleton classes involved in the meta-class. The location of center of meta-class in CCR is considered with the same (similar) distance to all the involved singleton classes' centers.
 Moreover, when the training data is not available, we have also proposed several credal clustering methods \cite{BCM,CCM,MECM} in different cases. Nevertheless, these previous credal classification methods are mainly for dealing with  complete pattern without taking into account the missing values.

 In our recent work, a prototype-based credal classification (PCC) \cite{PCC} method for the incomplete patterns has been introduced to capture the imprecise information caused by the missing values. The object hard to correctly classify are committed to a suitable meta-class by PCC, which well characterizes the imprecision of classification due to absence of part attributes and also reduces the misclassification errors. In PCC, the missing values in all the incomplete patterns are imputed using prototype of each class center, and the edited pattern with each imputation is respectively classified by a standard classifier (for complete pattern). With PCC, one obtains $c$ pieces of classification results for each incomplete pattern in a $c$ class problem, and the global fusion of the $c$ results is given for the credal classification. Unfortunately, PCC classifier is computationally greedy and time-consuming, and the imputation of missing values based on class prototype is not so precise. In order to overcome the limitations of PCC, we propose a new credal classification method for incomplete pattern with adaptive imputation of missing values, and it can be called Credal Classification with Adaptive Imputation (CCAI) for short.

The pattern to classify usually consists of multiple attributes. Sometimes, the class of the pattern can be precisely determined using only a part (a subset) of the available attributes, and it implies that the other attributes are redundant and in fact unnecessary for the classification. So a new method of credal classification with adaptive imputation strategy (i.e. CCAI) for missing values is proposed. In CCAI, we attempt to classify the object only using the known attributes value at first. If a specific classification result is obtained, it very likely means that the missing values are not very necessary for the classification, and we directly take the decision on the class of the object based on this result. However, if the object cannot be clearly classified with the available information, it indicates that the missing information included in the missing attribute values is probably very crucial for making the classification. In this case, we present a sophisticated classification strategy for the edition of pattern based on the proper imputation of missing values.

K-nearest neighbors-based imputation method usually provides pretty good performances for the estimation of missing values, but the its main drawback is the big computational burden. To reduce the computational burden, Self-Organizing Map (SOM) \cite{SOM} is applied in each class, and the optimized weighting vectors are used to represent the corresponding class. Then, the $K$ nearest weighting vectors of the object in each class are respectively employed to estimate the missing values. For the classification of original incomplete pattern (without imputation of missing values) or the edited pattern (with imputation of missing values), we adopt the ensemble classifier approach. One can respectively get the simple classification result according to each training class, and each classification result is represented by a simple basic belief assignment (BBA) including two focal elements (i.e. singleton class and ignorant class) only. The belief of the object belonging to each class is calculated based on the distance to the corresponding prototype, and the other belief is committed to the ignorant element. The fusion (ensemble) of these multiple BBA's is then used to determine the class of the object. If the object is directly classified using only the known values, Dempster-Shafer\footnote{Although the rule has been proposed originally by Arthur Dempster, we prefer to call it Dempster-Shafer rule because it has been widely promoted by Shafer in \cite{Shafer}.} (DS) fusion rule \cite{Shafer} is applied
 because of the simplicity of this rule and also because the BBA's to fuse are usually in low conflict. In this case, a specific result is obtained with DS rule. Otherwise, a new fusion rule inspired by Dubois and Prade (DP) rule \cite{DP} is used to classify the edited pattern with proper imputation of its missing values. Because the estimation of the missing values can be quite uncertain, it naturally induces an imprecise classification. So the partial conflicting beliefs will be kept and committed to the associated meta-classes in this new rule to reasonably reveal the potential imprecision of the classification result.

In this paper, we present an credal classification method with adaptive imputation of missing values based on belief function theory for dealing with the incomplete patterns, and it is organized as follows. The basics of belief function theory and Self-Organizing Map is briefly recalled in section \ref{secBK}.  The new credal classification method for incomplete patterns is presented in the section \ref{secCCI}, and the proposed method is then tested and evaluated in section \ref{secExamples} compared with several other classical methods. The paper is concluded in the final.

%======================================
\section{Background knowledge}
%======================================
\label{secBK}

Belief function theory (BFT) can well characterize the uncertain and imprecise information, and it is used in this work for the classification of patterns.
SOM technique is employed to find the optimized weighting vectors which are used to represent the corresponding class, and this can reduce the computation burden in the estimation of the missing values based on K-NN method. So the basic knowledge on BFT and SOM will be briefly recalled.

%===================================================
\subsection {Basis of belief function theory}
%====================================================
\label{secBF}
The Belief Function Theory (BFT) introduced by Glenn Shafer is also known as Dempster-Shafer Theory (DST), or the Mathematical Theory of Evidence \cite{Shafer,TBM1,Jean}. Let us consider a frame of discernment consisting of $c$ exclusive and exhaustive hypotheses (classes) denoted by $\Omega=\{\omega_i, i=1,2,\ldots,c\}$. The power-set of $\Omega$ denoted $2^\Omega$ is the set of all the subsets of $\Omega$, empty set included. For example, if $\Omega=\{\omega_1,\omega_2,\omega_3\}$, then $2^{\Omega}=\{\emptyset, \omega_1,\omega_2,\omega_3, \omega_1\cup \omega_2,\omega_1\cup \omega_3,\omega_2\cup \omega_3,\Omega\}$.
In the classification problem, the singleton element (e.g. $\omega_i$) represents a specific class. In this work, the disjunction (union) of several singleton elements is called a {\it{meta-class}} which characterizes the partial ignorance of classification. Examples of meta-classes are $\omega_i\cup \omega_j$, or $ \omega_i\cup \omega_j\cup \omega_k$. In BFT, one object can be associated with different singleton elements as well as with sets of elements according to a basic belief assignment (BBA), which is a function $m(.)$ from  $2^\Omega$ to $[0, 1]$ satisfying $ m(\emptyset)=0$ and  the normalization condition $ \sum\limits_{A\in 2^\Omega}m(A)=1 $. The subsets $A$ of $\Omega$ such that $m(A)>0$ are called the focal elements of the belief mass $m(.)$.

The credal classification (or partitioning) \cite{ECM} is defined as $n$-tuple $M=(\mathbf{m_1},\cdots,\mathbf{m_n})$ of BBA's, where $\mathbf{m}_i$ is the basic belief assignment of the object $\mathbf{x}_i \in X$, $i=1,\ldots, n$ associated with the different elements in the power-set $2^{\Theta}$. The credal classification allows the objects to belong to the specific classes and  the sets of classes corresponding to meta-classes with different belief mass assignments. The credal classification can well model the imprecise and uncertain information thanks to the introduction of meta-class.

For combining multiple sources of evidence represented by a set of BBA's, the well-known Dempster's rule \cite{Shafer} is still widely used, even if its justification is an open debate and questionable in the community \cite{Zadeh79,IJIS2014}. The  combination of two BBA's $m_1(.)$ and $m_2(.)$ over $2^\Omega$  is done with DS rule of combination defined by $m_{DS}(\emptyset)=0$ and for $A\neq \emptyset, B, C \in 2^\Omega$ by

 \begin{align}
 m_{DS}(A)=\frac{\sum \limits_{B \cap C=A}m_1(B) m_2(C)}{1-\sum \limits_{B\cap C = \emptyset}m_1(B)m_2(C)}
 \label{eq:DS}
 \end{align}
DS rule is commutative and associative, and makes a compromise between the specificity and complexity for the combination of BBA's. With this rule, all the conflicting beliefs $\sum \limits_{B\cap C = \emptyset}m_1(B)m_2(C)$ are proportionally redistributed back to the focal elements through a classical normalization step. However, this redistribution can yield unreasonable results in the high conflicting cases \cite{Zadeh79}, as well as in some special low conflicting cases as well \cite{IJIS2014}. That is why different rules of combination have emerged to overcome its limitations. Among the possible alternatives of DS rule, we find Smets' conjunctive rule (used in  his transferable belief model (TBM) \cite{TBM1}), Dubois-Prade (DP) rule \cite{DP}, and more recently the more complex Proportional Conflict Redistributions (PCR) rules \cite{Fusion2005}. Unfortunately, DP and PCR rules are less appealing from implementation standpoint since they are not associative, and they become complex to use when more than two BBA's have to be combined altogether.

%===================================================
\subsection {Overview of Self-Organizing Map}
%====================================================
Self-Organizing Map (SOM) (also called Kohonen map) \cite{SOM} introduced by Teuvo Kohonen is a type of artificial neural network (ANN),
 and it is trained by unsupervised learning method. SOM defines a mapping from the input space to a low-dimensional (typically two-dimensional) grid of $M\times N$ nodes. So it allows to approximate the feature space dimension (e.g. a real input vector
 $\mathbf{x}\in \mathbb{R}^p$) into a projected 2D space, and it is still able to preserve the topological properties of the input space using a neighborhood function. Thus, SOM is very useful for visualizing low-dimensional views of high-dimensional data by a non linear projection.

The node at position $(i, j), i=1, \dots M, j=1,\ldots, N$ corresponds to
a weighting vector denoted by $\sigma(i,j)\in \mathbb{R}^p$. An input vector
 $\mathbf{x}\in \mathbb{R}^p$ is to be compared to each $\sigma(i, j)$, and the neuron whose weighting vector is the most close (similar) to $\mathbf{x}$ according to a given metric is called the best matching unit (BMU), which is defined as the output of SOM with respect to $\mathbf{x}$.
 In real applications, the Euclidean distance is usually used to compare  $\mathbf{x}$
and $\sigma(i,j)$.
 The input pattern $\mathbf{x}$ can be mapped onto the SOM at location $(i, j)$
where $\sigma(i,j)$ is with the minimal distance to $\mathbf{x}$.
It is considered that the SOM achieves a non-uniform quantization that
transforms $\mathbf{x}$ to $\sigma_\mathbf{x}$ by minimizing the given metric (e.g. distance measure) \cite{SOM2014}.

In SOM, the competitive learning is adopted, and the
training algorithm is iterative. The initial values of the weighting vectors $\mathbf{\sigma}$ may be set randomly, but they will
converge to a stable value at the end of the training process. When an input vector is fed to the network, its Euclidean distance to all weight vectors is computed. Then the BMU whose weight vector is most similar to the input vector is found, and the weights of the BMU and neurons close to it in the SOM grid are adjusted towards the input vector. The magnitude of the change decreases with time and with distance (within the grid) from the BMU. The detailed information about SOM can be found in \cite{SOM}.

In this work, SOM is applied in each training class to obtain the optimized weighting vectors that are used to represent the corresponding class. The number of the weighting vectors is much smaller than the original samples in the associated training class. We will utilize these weighting vectors rather than the original samples to estimate the missing values in the object (incomplete pattern), and this could effectively reduce the computation burden.

%==========================================================
\section{Credal classification of incomplete pattern}
%==========================================================
\label{secCCI}

Our new method consists of two main steps. In the first step, the object (incomplete pattern) is directly classified according to the known attribute values only, and the missing values are ignored. If one can get a specific classification result, the classification procedure is done because the available attribute information is sufficient for making the classification. But if the class of the object cannot be clearly identified in the first step, it means that the unavailable information included in the missing values is likely crucial for the classification. In this case, one has to enter in the second step of the method to classify the object with a proper imputation of missing values.
 In the classification procedure, the original or edited pattern will be respectively classified according to each class of training data. The global fusion of these classification results, which can be considered as multiple sources of evidence represented by BBA's, is then used for the credal classification of the object.
Our new method for credal classification of incomplete pattern with adaptive imputation of missing values is referred as Credal Classification with Adaptive Imputation, or just as CCAI for conciseness.  CCAI is based on belief function theory, which can well manage the uncertain and imprecise information caused by the missing values in the classification.

%==================================================================================
\subsection{First step: Direct classification of incomplete pattern using the available data}
%==================================================================================
\label{secDc}
Let us consider a set of test patterns (samples) $X=\{\mathbf{x}_1, \ldots, \mathbf{x}_n\}$ to be classified based on a set of labeled training patterns $Y=\{\mathbf{y}_1, \ldots, \mathbf{y}_s\}$ over the frame of discernment $\Omega=\{\omega_1,\ldots, \omega_c\}$.
In this work, we focus on the classification of incomplete pattern in which some attribute values are absent. So we consider all the test patterns (e.g.  $\mathbf{x}_i, i=1,\ldots, n$) with several missing values. The training data set $Y$ may also have incomplete patterns in some applications. However, if the incomplete patterns take a very small amount say less than 5\% in the training data set, they can be ignored in the classification. If the percentage of incomplete patterns is big, the missing values must usually be estimated at first, and the classifier will be trained using the edited (complete) patterns. In the real applications, one can also just choose the complete labeled patterns to include in the training data set when the training information is sufficient. So for simplicity and convenience, we consider that the labeled samples (e.g.  $\mathbf{y}_j, j=1,\ldots, s$) of the training set $Y$ are all complete patterns in the sequel.

In the first step of classification, the incomplete pattern say $\mathbf{x}_i$ will be respectively classified according to each training class by a normal classifier (for dealing with the complete pattern) at first, and all the missing values are ignored here.  In this work, we adopt a very simple classification method\footnote{Many other normal classifiers (e.g. K-NN) can be selected here depending on the preference of user, and we propose to use this simple classification method because of its low computation complexity.} for the convenience of computation, and $\mathbf{x}_i$ is directly classified based on the distance to the prototype of each class.

 The prototype of each class $\{\mathbf{o}_1, \ldots, \mathbf{o}_c\}$ corresponding to $\{\omega_1,\ldots, \omega_c\}$ is given by the arithmetic average vector of the training patterns in the same class. Mathematically, the prototype is computed for $g=1,\dots, c$ by
\begin{equation}
\mathbf{o}_g=\frac{1}{N_g}\sum\limits_{\mathbf{y}_j\in \omega_g}\mathbf{y}_j
\label{eq:prototype}
\end{equation}
\noindent
where $N_g$ is the number of the training samples in the class $\omega_g$.

In a $c$-class problem, one can get $c$ pieces of simple classification result for $\mathbf{x}_i$ according to each class of training data, and each result is represented by a simple BBA's including two focal elements, i.e. the singleton class and the ignorant class ($\Omega$) to characterize the full ignorance.
 The belief of $\mathbf{x}_i$ belonging to class $\omega_g$ is computed based on the distance between $\mathbf{x}_i $ and the corresponding prototype $\mathbf{o}_g$. Normalized Euclidean distance as eq. \eqref{eq:dist} is adopted here to deal with the anisotropic class, and the missing values are ignored in the calculation of this distance. The other mass of belief is assigned to the ignorant class $\Omega$. Therefore, the BBA's construction is done by
\begin{equation}
\begin{cases}
m_i^{\mathbf{o}_g}(\omega_g)=e^{-\eta d_{ig}}\\
m_i^{\mathbf{o}_g}(\Omega)=1-e^{-\eta d_{ig}}
\end{cases}
\label{eq:m}
\end{equation}
\noindent
with
\begin{equation}
d_{ig}=\sqrt{\frac{1}{p} \sum_{j=1}^p {\left(\frac{x_{ij}-o_{gj}}{\delta_{gj}}\right)}^2}
\label{eq:dist}
\end{equation}
\noindent
and
\begin{equation}
\delta_{gj}=\sqrt{\frac{1}{N_g}\sum_{\mathbf{y}_i\in \omega_g}{(y_{ij}-o_{gj})}^2}
\label{eq:delta}
\end{equation}
\noindent
 where $x_{ij}$ is value of $\mathbf{x}_i$ in $j$-th dimension, and $y_{ij}$ is value of $\mathbf{y}_i$ in $j$-th dimension.
$p$ is the number of available attribute values in the object $\mathbf{x}_i$. The coefficient $1/p$ is necessary to normalize the distance value because each test sample can have a different number of missing values. $\delta_{gj}$ is the average distance of all training samples in class $\omega_g$ to the prototype $o_g$ in $j$-th dimension. $N_g$ is the number of training samples in $\omega_g$. $\eta$ is a tuning parameter, and the bigger $\eta$ generally yields smaller mass of belief on the specific class $w_g$. It is usually recommended to take $\eta\in [0.5, 0.8]$ according to our various tests, and $\eta=0.7$ can be considered as default value.

Obviously, the smaller distance measure, the bigger mass of belief on the singleton class. This particular structure of BBA's indicates that we can just confirm the degree of the object $\mathbf{x}_i$ associated with the specific class $\omega_g$ only according to training data in $\omega_g$. The other mass of belief reflects the level of belief one has on full ignorance, and it is committed to the ignorant class $\Omega$. Similarly, one calculates $c$ independent BBA's $m_i^{\mathbf{o}_g}(\omega_g), g=1,\ldots, c$ based on the different training classes.

Before combining these $c$ BBA's, we examine whether a specific classification result can be derived from these $c$ BBA's. This is done as follows:
if it holds that $m_i^{\mathbf{o}_{1st}}(\omega_{1st})={argmax}_{g}(m_i^{\mathbf{o}_g}(\omega_g))$, then the object will be considered to belong very likely to the class $\omega_{1st}$, which obtains the biggest mass of belief in the $c$ BBA's. The class with the second biggest mass of belief is denoted $\omega_{2nd}$.

The distinguishability degree $\chi_i\in (0, 1]$ of an object $\mathbf{x}_i$ associated with different classes is defined by:
\begin{equation}
\chi_i=\frac{m_i^{\mathbf{o}_{2nd}}(\omega_{2nd})}{m_i^{\mathbf{o}_{max}}(\omega_{max})}
\end{equation}

Let $\epsilon$ be a chosen small positive distinguishability threshold value in $(0, 1]$. If the condition $\chi_i\leq \epsilon$ is satisfied, it means that all the classes involved in the computation of $\chi_i$ can be clearly distinguished of $\mathbf{x}_i$. In this case, it is very likely to obtain a specific classification result from the fusion of the $c$ BBA's. The condition  $\chi_i\leq \epsilon$ also indicates that the available attribute information is sufficient for making the classification of the object, and the imputation of the missing values is not necessary. If $\chi_i\leq \epsilon$ condition holds, the  $c$ BBA's are directly combined with DS rule to obtain the final classification results of the object because DS rule usually produces specific combination result with acceptable computation burden in the low conflicting case. In such case, the meta-class is not included in the fusion result, because these different classes are considered distinguishable based on the condition of distinguishability. Moreover, the mass of belief of the full ignorance class $\Omega$, which represents the noisy data (outliers), can be proportionally redistributed to other singleton classes for more specific results if one knows a priori that the noisy data is not involved.

If the distinguishability condition  $\chi_i\leq \epsilon$ is not satisfied, it means that the classes $\omega_{1st}$ and $\omega_{2nd}$ cannot be clearly distinguished for the object with respect to the chosen threshold value $\epsilon$, indicating that missing attribute values play almost surely a crucial role in the classification. In this case, the missing values must be properly imputed to recover the unavailable attribute information before entering the classification procedure. This is the Step 2 of our method which is explained in the next subsection.

%====================================================================
\subsection{Second step: Classification of incomplete pattern with imputation of missing values}
%====================================================================

%=====================================================
 \subsubsection{Multiple estimation of missing values}
 %====================================================

 In the estimation of the missing attribute values, there exist various methods. Particularly, the K-NN imputation method generally provides good performance. However,
 the main drawback of KNN method is its big computational burden, since one needs  to calculate the distances of the object with all the training samples.
Inspired by \cite{SOM2014}, we propose to use the Self Organized Map (SOM) technique \cite{SOM} to reduce the computational complexity. SOM can be applied in each class of training data, and then $M\times N$ weighting vectors will be obtained after the optimization procedure. These optimized  weighting vectors allow to characterize well the topological features of the whole class, and they will be used to represent the corresponding data class. The number of the  weighting vectors is usually small (e.g. $5\times 6$). So the $K$ nearest neighbors of the test pattern associated with these weighting vectors in the SOM can be easily found with low computational complexity\footnote{ The training of SOM using the labeled patterns becomes time consuming when the number of labeled patterns is big, but fortunately it can be done off-line. In our experiments, the running time performance shown in the results doesn't include the computational time spent for the off-line procedures. }. The selected weighting vector no. $k$ in the class $\omega_g$, $g=1,\ldots, c$ is denoted $\mathbf{\sigma}_{k}^{\omega_g}$, for $k=1,\ldots, K$.

In each class, the $K$ selected close  weighting vectors provide different contributions (weight) in the estimation of missing values, and the weight $p_{ik}^{\omega_g}$ of each vector is defined based on the distance between the object  $\mathbf{x}_i$ and weighting vector $\mathbf{\sigma}_{k}^{\omega_g}$.
\begin{equation}
p_{ik}^{\omega_g}=e^{(-\lambda d_{ik}^{\omega_g})}
\end{equation}
\noindent
with
\begin{equation}
\lambda=\frac{cNM(cNM-1)}{2\sum\limits_{i,j}d(\mathbf{\sigma}_{i},\mathbf{\sigma}_{j})}
\end{equation}
where  $d_{ik}^{\omega_g}$ is the Euclidean distance between $\mathbf{x}_i$ and the neighbor $\mathbf{o}_k^{\omega_g}$ ignoring the missing values, and  $\frac{1}{\lambda}$ is the average distance between each pair of  weighting vectors produced by SOM in all the classes; $c$ is the number of classes; $M\times N$ is the number of  weighting vectors obtained by SOM in each class; and $d(\mathbf{\sigma}_{i},\mathbf{\sigma}_{j})$ is the Euclidean distance between any two weighting vectors $\mathbf{\sigma}_{i}$ and $\mathbf{\sigma}_{j}$.

The weighted mean value $\hat{\mathbf{y}}_i^{\omega_g}$ of the selected $K$  weighting vectors in class training class $\omega_g$ will be used for the imputation of missing values. It is calculated by
\begin{equation}
\hat{\mathbf{y}}_i^{\omega_g}=\frac{\sum\limits_{k=1}^K p_{ik}^{\omega_g}\mathbf{\sigma}_{k}^{\omega_g}}{\sum\limits_{k=1}^K p_{ik}^{\omega_g}}
\end{equation}

The missing values in $\mathbf{x}_i$ will be filled by the values of $\hat{\mathbf{y}}_i^{\omega_g}$ in the same dimensions. By doing this, we get the edited pattern $\mathbf{x}_i^{\omega_g}$ according to the training class $\omega_g$.

Then $\mathbf{x}_i^{\omega_g}$ will be simply classified only based on the training data in $\omega_g$ as similarly done in the direct classification of incomplete pattern using eq. \eqref{eq:m} of Step 1 for convenience\footnote{Of course, some other sophisticated classifiers can also be applied here according to the selection of user, but the choice of classifier is not the main purpose of this work.}.

The classification of $\mathbf{x}_i$ with the estimation of missing values is also respectively done based on the other training classes according to this procedure.  For a $c$-class problem, there are $c$ training classes, and therefore one can get $c$ pieces of classification results with respect to one object.

%===========================================================
\subsubsection{Ensemble classifier for credal classification}
%===========================================================

These $c$ pieces of results obtained by each class of training data in a $c$-class problem are considered with different weights, since the estimations of the missing values according to different classes have different reliabilities.
 The weighting factor of the classification result associated with the class $w_g$ can be defined by the sum of the weights of the $K$ selected SOM  weighting vectors for the contributions to the missing values imputation in $\omega_g$, which is given by
\begin{equation}
 \rho_i^{\omega_g}=\sum\limits_{k=1}^K p_{ik}^{\omega_g}
 \end{equation}

The result with the biggest weighting factor $\rho_i^{\omega_{max}}$ is considered as the most reliable, because one assumes that the object must belong to one of the labeled classes (i.e. $w_g$, $g=1,\ldots,c$). So the biggest weighting factor will be normalized as one. The other relative weighting factors are defined by:
\begin{equation}
 \hat{\alpha}_i^{\omega_g}=\frac{\rho_i^{\omega_g}}{\rho_i^{\omega_{max}}}
 \label{eq:alpha1}
 \end{equation}

If the condition\footnote{ The threshold $\epsilon$ is the same as in section \ref{secDc}, because it is also used to measure the distinguishability degree here.}  $\hat{\alpha}_i^{\omega_g}<\epsilon$  is satisfied, the corresponding estimation of the missing values and the classification result are not very reliable. Very likely, the object does not belong to this class. It is implicitly assumed that the object can belong to only one class in reality. If this result whose relative weighting factor is very small (w.r.t. $\epsilon$) is still considered useful, it will be (more or less) harmful for the final classification of the object. So if the condition $\hat{\alpha}_i^{w_g}<\epsilon$ holds, then the relative weighting factor is set to zero. More precisely, we will take
\begin{equation}
  \alpha_i^{\omega_g}=
  \begin{cases}
  0, &\quad \text{if} \quad \hat{\alpha}_i^{\omega_g}<\epsilon\\
  \frac{\rho_i^{\omega_g}}{\rho_i^{\omega_{max}}}, & \quad  \text{otherwise.}
  \end{cases}
  \label{eq:alpha2}
  \end{equation}

After the estimation of weighting (discounting) factors $\alpha_i^{\omega_g}$, the $c$ classification results (the BBA's $m_i^{\mathbf{o}_g}(.)$) are classically discounted \cite{Shafer} by
\begin{equation}
\begin{cases}
\hat{m}_i^{\mathbf{o}_g}(\omega_g)=\alpha_i^{\omega_g} m_i^{\mathbf{o}_g}(\omega_g) \\
\hat{m}_i^{\mathbf{o}_g}(\Omega)=1-\alpha_i^{\omega_g}+\alpha_i^{\omega_g} m_i^{\mathbf{o}_g}(\Omega)
\label{eq:discount1}
\end{cases}
\end{equation}

These discounted BBA's will be globally combined to get the credal classification result. If $\alpha_i^{\omega_g}=0$, one gets $\hat{m}_i^{\mathbf{o}_g}(\Omega)=1$, and this fully ignorant (vacuous) BBA plays a neutral role in the global fusion process for the final classification of the object.

Although we have done our best to estimate the missing values, the estimation can be quite imprecise when the estimations are obtained from different class with the similar weighting factors, and the different estimations probably lead to distinct classification results. In such case, we prefer to cautiously keep (rather to ignore) the uncertainty, and maintain the uncertainty in the classification result. Such uncertainty can be well reflected by the conflict of these classification results represented by the BBA's.
DS rule is not suitable here, because all the conflicting beliefs are distributed to other focal elements. A particular combination rule inspired by DP rule is introduced here to fuse these BBA's according to the current context. In our new rule, the partial conflicting beliefs are prudently transferred to the proper meta-class to reveal the imprecision degree of the classification caused by the missing values. This new rule of combination is defined by:
\begin{equation}
\begin{cases}
m_i(\omega_g)=\hat{m}_i^{\mathbf{o}_g}(\omega_g)\prod\limits_{j\neq g} \hat{m}_i^{\mathbf{o}_j}(\Omega)\\
m_i(A)=\prod\limits_{\bigcup\limits_{j}\omega_j=A} \hat{m}_i^{\mathbf{o}_j}(\omega_j)\prod\limits_{k\neq j}\hat{m}_i^{\mathbf{o}_k}(\Omega)
\end{cases}
\end{equation}

The test pattern can be classified according to the fusion results, and the object is considered belonging to the class (singleton class or meta-class) with the maximum mass of belief. This is called hard credal classification. If one object is classified to a particular class, it means that this object has been correctly classified with the proper imputation of missing values. If one object is committed to a meta-class (e.g. $A\cup B$), it means that we just know that this object belongs to one of the specific classes (e.g. $A$ or $B$) included in the meta-class, but we cannot specify which one. This case can happen when the missing values are essential for the accurate classification of this object, but the missing values cannot be estimated very well according to the context, and different estimations will induce the classification of the object into distinct classes (e.g. $A$ or $B$).

\begin{figure}[!thbt]
 \begin{center}
 	\includegraphics[width=1\linewidth]{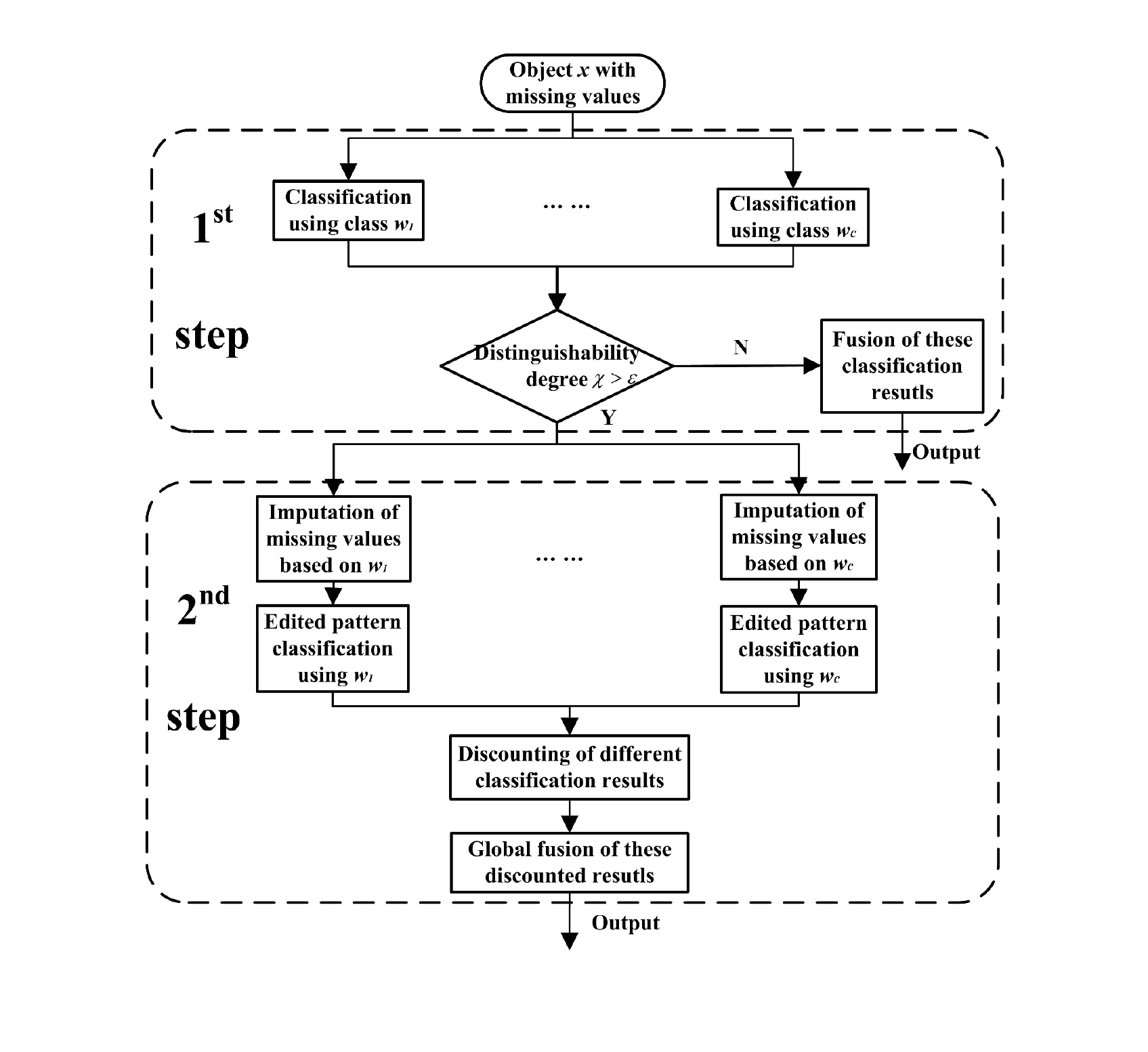}
		\hfill
			\caption{Flowchart of the proposed CCAI method. \label{figintro}}
\label{flowchart}
\end{center}
\end{figure}

For convenience,  Fig. \ref{flowchart} shows the functional flowchart of this new CCAI method.

\vspace{5mm}
\noindent
{\bf{Guideline for tuning of the parameters $\epsilon$ and $\eta$}}: The tuning of parameters $\eta$ and $\epsilon$ is very important in the application of CCAI.
$\eta$ in eq. \eqref{eq:m} is associated with the calculation of mass of belief on the specific class, and the bigger $\eta$ value will lead to smaller mass of belief committed to the specific class. Based on our various tests, we advise to take $\eta\in [0.5, 0.8]$, and the value $\eta=0.7$ can be taken as the default value.
The parameter $\epsilon$ is the threshold to tune for changing the classification strategy. It is also used in Eq. \eqref{eq:alpha2} for the calculation of the discounting factor. The bigger $\epsilon$ will make fewer objects going to the sophisticated classification procedure with the imputation of missing values, and it also forces more discounting factors to zero according to Eq. \eqref{eq:alpha2}, which implies that fewer simple classification results obtained based on each class can be useful in the global fusion step. So the bigger $\epsilon$ will makes fewer objects committed to the meta-classes (corresponding to the low imprecision of classification), but it increases the risk of misclassification error.  $\epsilon$ should be tuned according to the compromise one can accept between the misclassification error and imprecision (non specificity of classification decision). One can also apply the cross validation \cite{kfold} (e.g. leave-one-out method) in the training data space to find a suitable threshold, and the missing values in the test samples are randomly distributed in all the dimensions.

%%=============================
\section{Experiments}
%==============================
\label{secExamples}

Three experiments with artificial and real data sets have been used to test the performance of this new CCAI method compared with the K-NN imputation (KNNI) method \cite{knnI}, FCM imputation (FCMI) method \cite{FCMI1, FCMI2}, SOM imputation (SOMI) \cite{SOMI} method and our previous credal classification PCC method \cite{PCC}. SOM technique is also employed in the second step of CCAI method, but CCAI is different from the previous SOMI method. In SOMI method, SOM is applied for the whole training data set, and the missing values are precisely estimated based on an activation group composed of the best match node (unit) of input pattern and its close neighbors. Then, the edited pattern with the imputation of missing values can be classified using a standard classifier. Nevertheless, SOM is not involved in the first step of CCAI, and the object is directly classified ignoring the missing values. In the second step of CCAI, SOM is respectively applied in each training class, and multiple estimations of missing values can be obtained based on the input pattern's $K$ nearest weighting vectors corresponding to nodes of SOM in each class. Then different classification results will be produced according to different estimations, and these results are globally fused for final classification. The conflicting information committed to the meta-class is kept in the fusion to characterize the imprecision of classification in CCAI, but this cannot be done in SOMI.
These different methods have been programmed and tested with Matlab$\texttrademark$ software.

The evidential neural network classifier (ENN) \cite{ANNbf} is adopted in the sequel experiments to classify the edited pattern with the estimated values in PCC, KNNI and FCMI, since ENN produce generally good results in the classification\footnote{Other traditional classifiers for complete pattern can also be selected here according to the actual application.}. The evidential K-nearest neighbor (EK-NN) method \cite{eknn} is also used to classify the edited pattern in Experiment 3 with real data for comparison.
 The parameters of ENN and EK-NN can be automatically optimized as explained in \cite{ANNbf} and \cite{knnoptimi}. In SOMI, we use the $M\times N=6\times 8$ nodes for mapping the whole input data set consisting of all the training classes to the 2-dimensional grid, and it has good performance. In the applications of PCC, the tuning parameter $\epsilon$ can be tuned according to the imprecision rate one can accept. In CCAI, a small number of the nodes in the 2-dimensional grid of SOM is given by $M\times N=3\times 4$ for each class, and we take the value of $K=N=4$ in K-NN for the imputation of missing values. This seems to provide good result in the sequel experiments. In order to show the ability of CCAI and PCC to deal with the meta-classes, the hard credal classification is applied, and the class of each object is decided according to the criterion of the maximal mass of belief.

In our simulations, the misclassification is declared (counted) for one object truly originated from $\omega_i$  if it is classified into $A$ with $\omega_i\cap A = \emptyset$. If $\omega_i\cap A \neq \emptyset$ and $A\neq \omega_i$ then it will be considered as an imprecise classification.
  The error rate denoted by $Re$ is calculated by $Re=N_e/T$, where $N_e$ is number of misclassification errors, and $T$ is the number of objects under test.
  The imprecision rate denoted by $Ri_j$ is calculated by $Ri_j=Ni_j/{T}$, where $Ni_j$ is number of objects committed to the meta-classes with the cardinality value $j$. In our experiments, the classification of object is generally uncertain (imprecise) among a very small number (e.g. 2) of classes, and we only take $Ri_2$ here since there is no object committed to the meta-class including three or more specific classes.

 %===========================
\subsection{Experiment 1 (artificial data set)}
\begin{figure}[!thbt]
 \begin{center}
 	\includegraphics[width=1\linewidth]{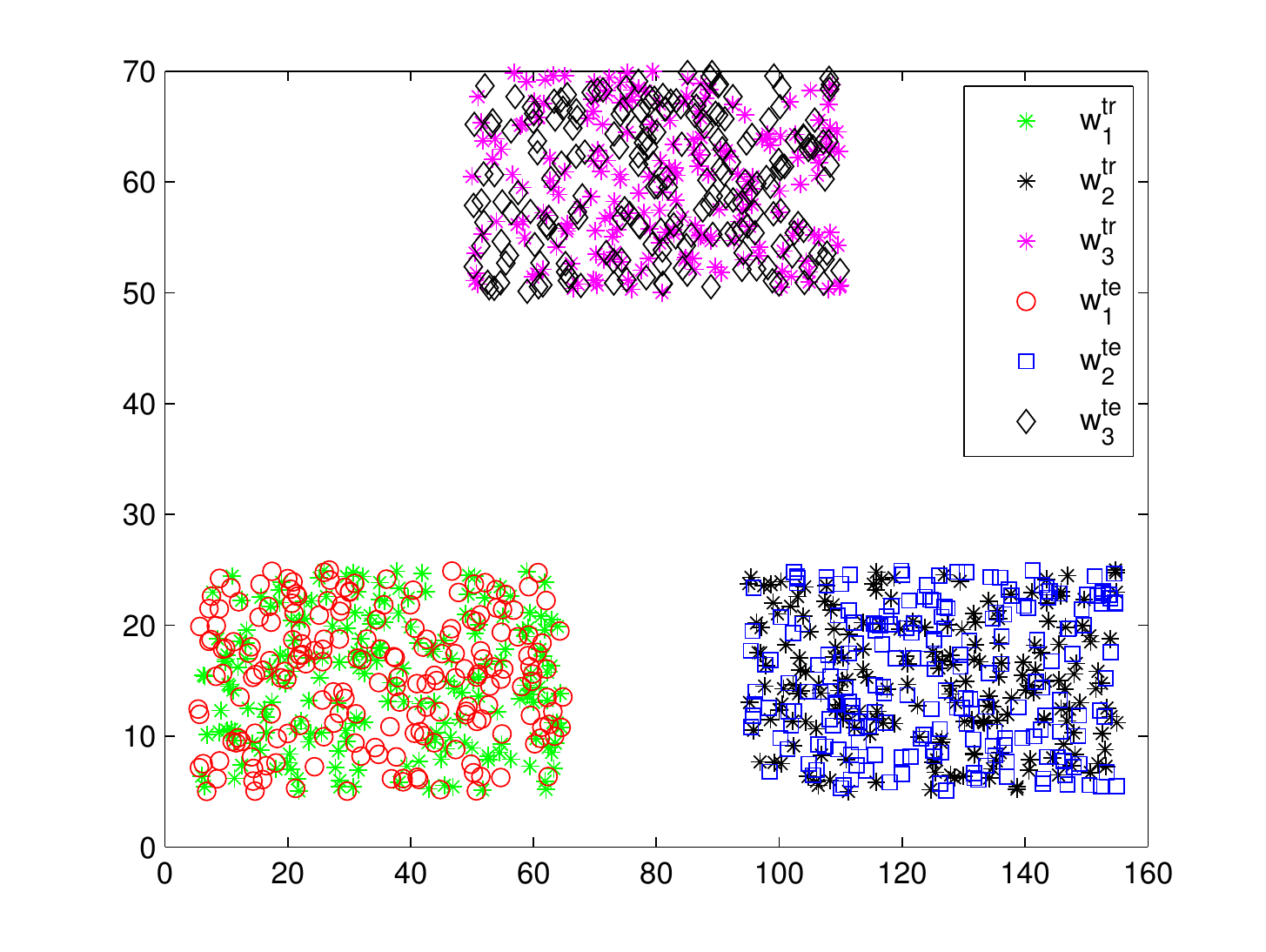}
		\hfill
			\caption{Training data and test data. \label{figdata}}
 \end{center}
\end{figure}

%============================
In the first experiment, we show the interest of credal classification based on belief functions with respect to the traditional classification working with probability framework. A 3-class data set $\Omega=\{\omega_1,\omega_2, \omega_3 \}$ obtained from three 2-D uniform distributions shown by Fig. \ref{figdata} is considered here. Each class has 200 training samples and 200 test samples, and there are 600 training samples and 600 test samples in total.
 The uniform distributions of the three classes are characterized by the following interval bounds:
\begin{center}
\begin{tabular}{c|cc}
& x-label interval & y-label interval\\
\hline
$\omega_1$&  (5, 65) & (5, 25)\\
$\omega_2$&  (95, 155)& (5,  25)\\
$\omega_3$&  (50, 110)& (50,  70)\\
\end{tabular}
\end{center}

  The values in the second dimension corresponding to y-coordinate of test samples are all missing. So test samples are classified according to the only one available value in the first dimension corresponding to x-coordinate.

\begin{center}
\begin{figure}[!thbt]
	\includegraphics[width=0.45\linewidth]{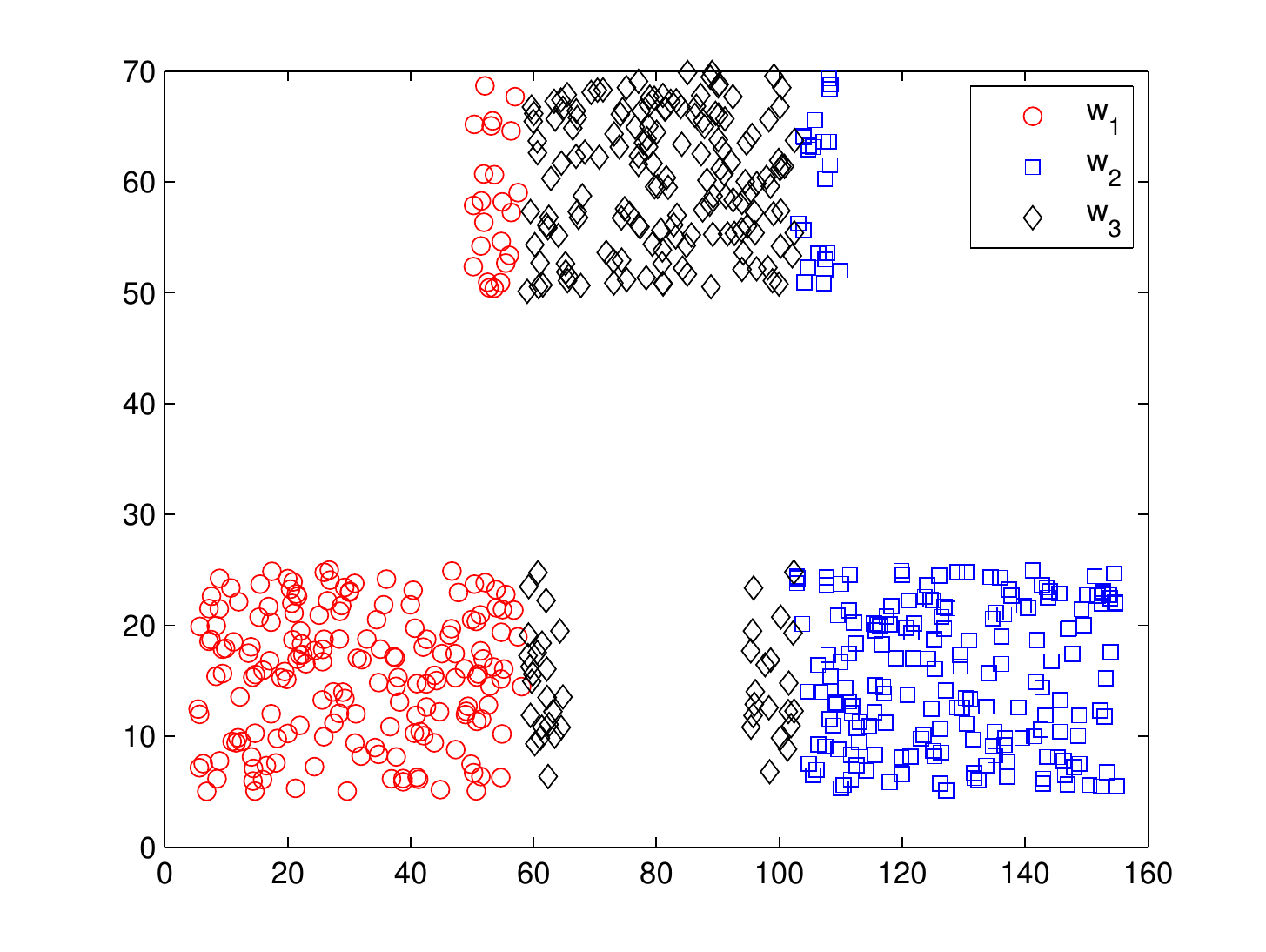}
	\hfill
	\includegraphics[width=.45\linewidth]{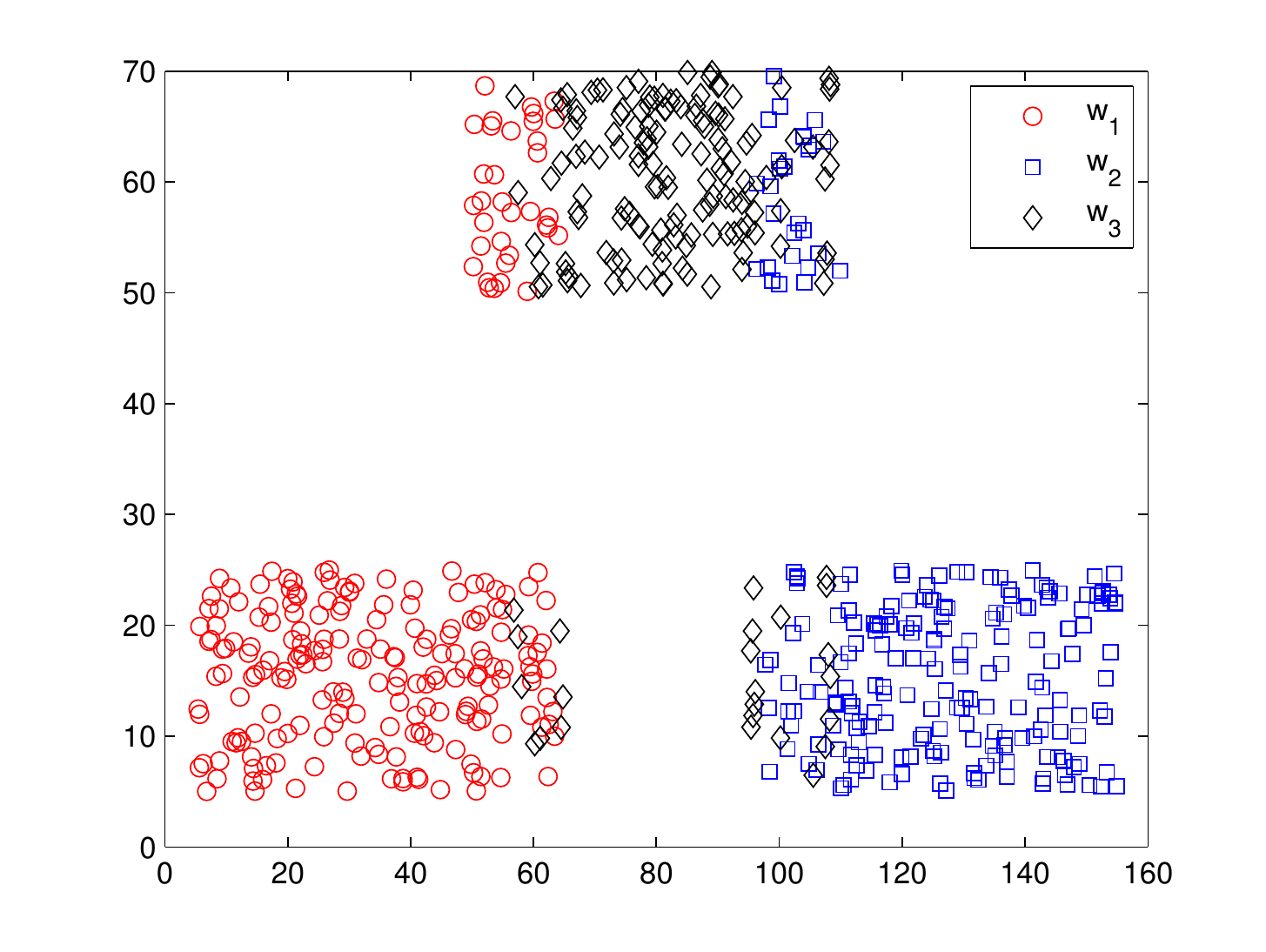}
	\hfill
\parbox{.450\linewidth}{\centering\small (a). Classification result by FCMI ($Re=14.67, time=0.0469s$).}
	\hfill
	\parbox{.450\linewidth}{\centering\small (b).  Classification result by KNNI ($Re=14.17, time=7.9531s$).  }
\hfill
	\includegraphics[width=0.45\linewidth]{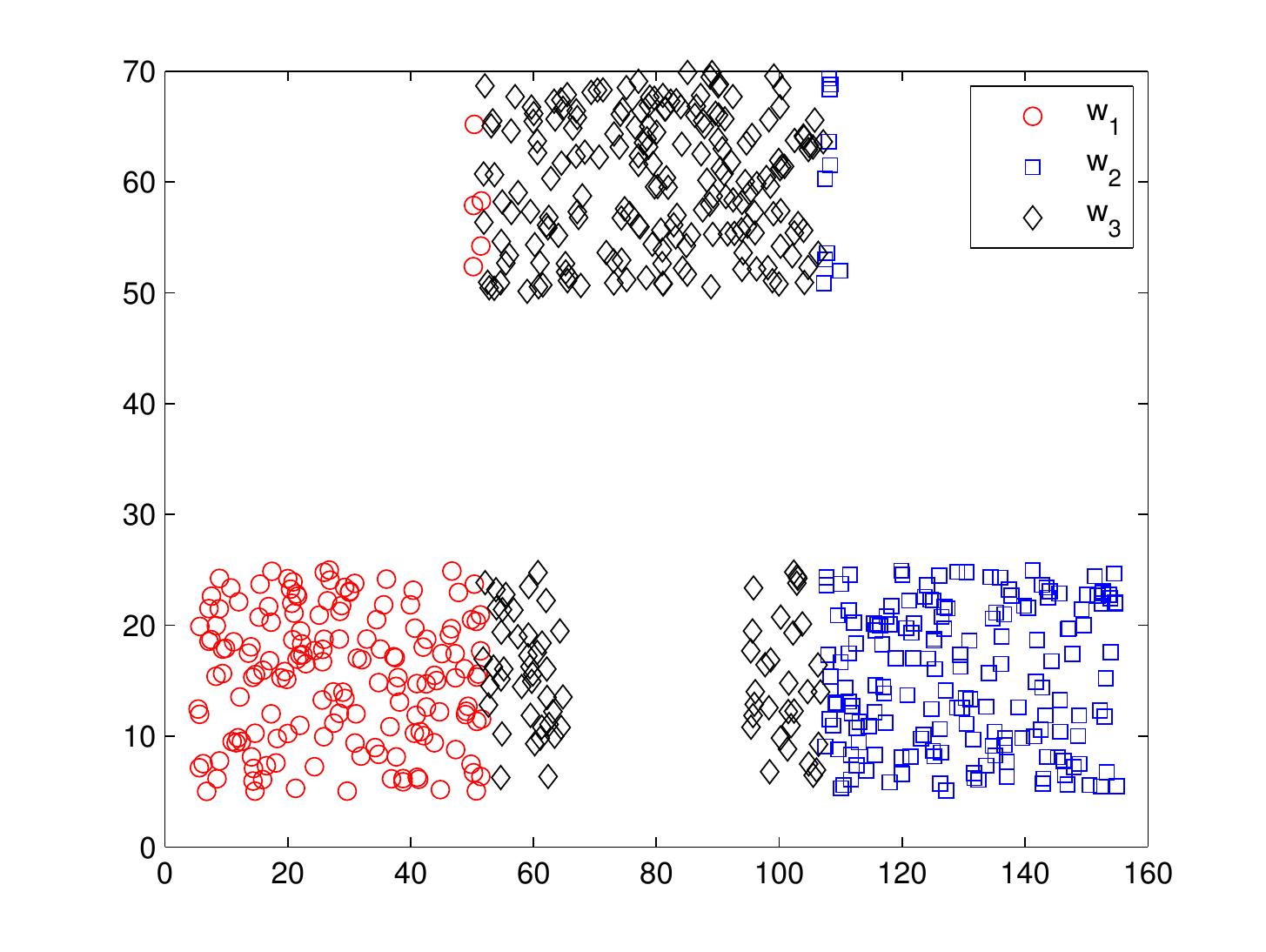}
	\hfill
	\includegraphics[width=0.45\linewidth]{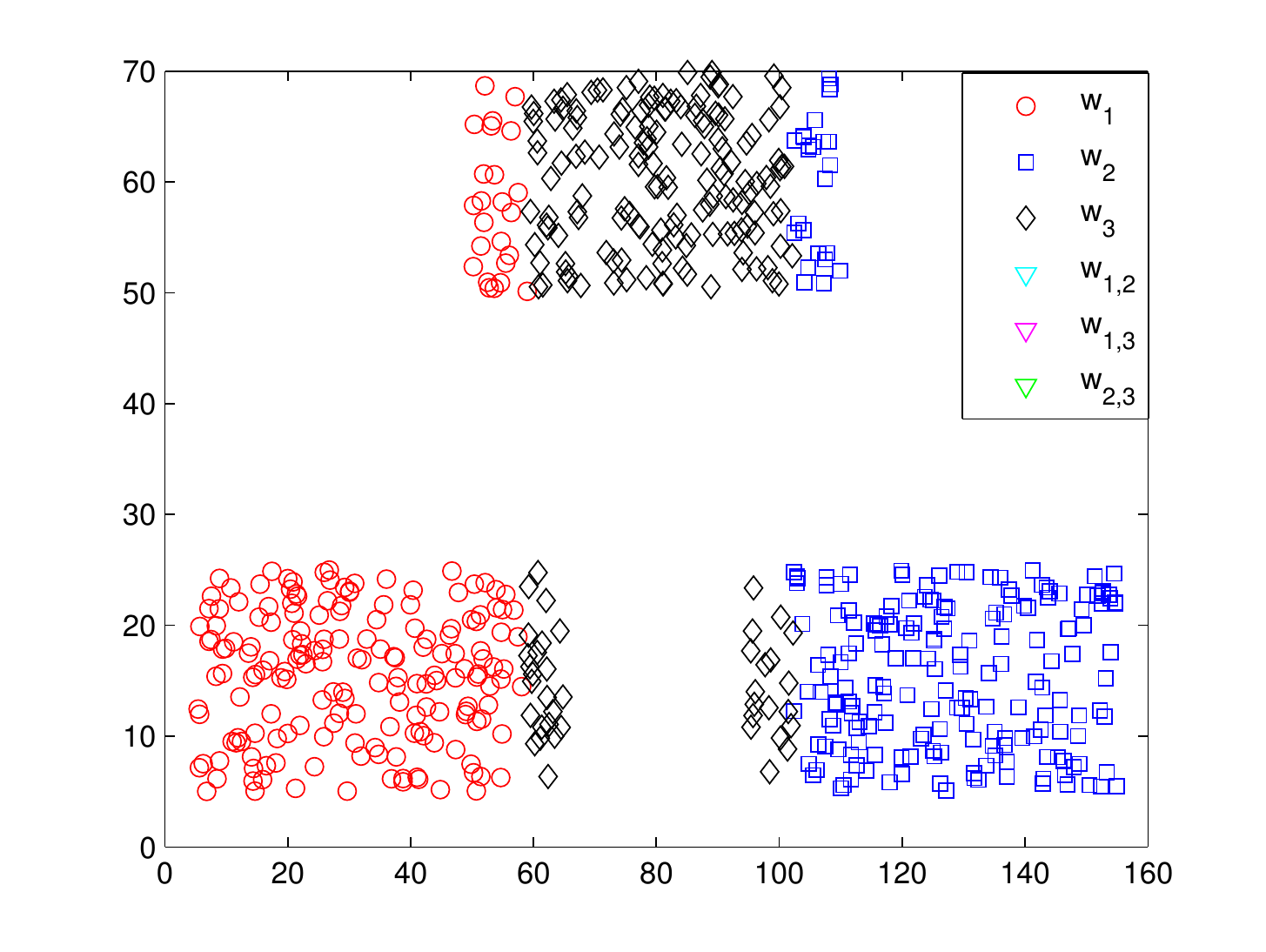}
	\hfill
\parbox{.450\linewidth}{\centering\small (c). Classification result by SOMI ($Re=14.33, time=0.9063s$).}
	\hfill
	\parbox{.450\linewidth}{\centering\small (d). Classification result only by $1^{st}$ step of SCCAI ($Re=14.83, time=0.0156s$). }
	\hfill
 \includegraphics[width=0.450\linewidth]{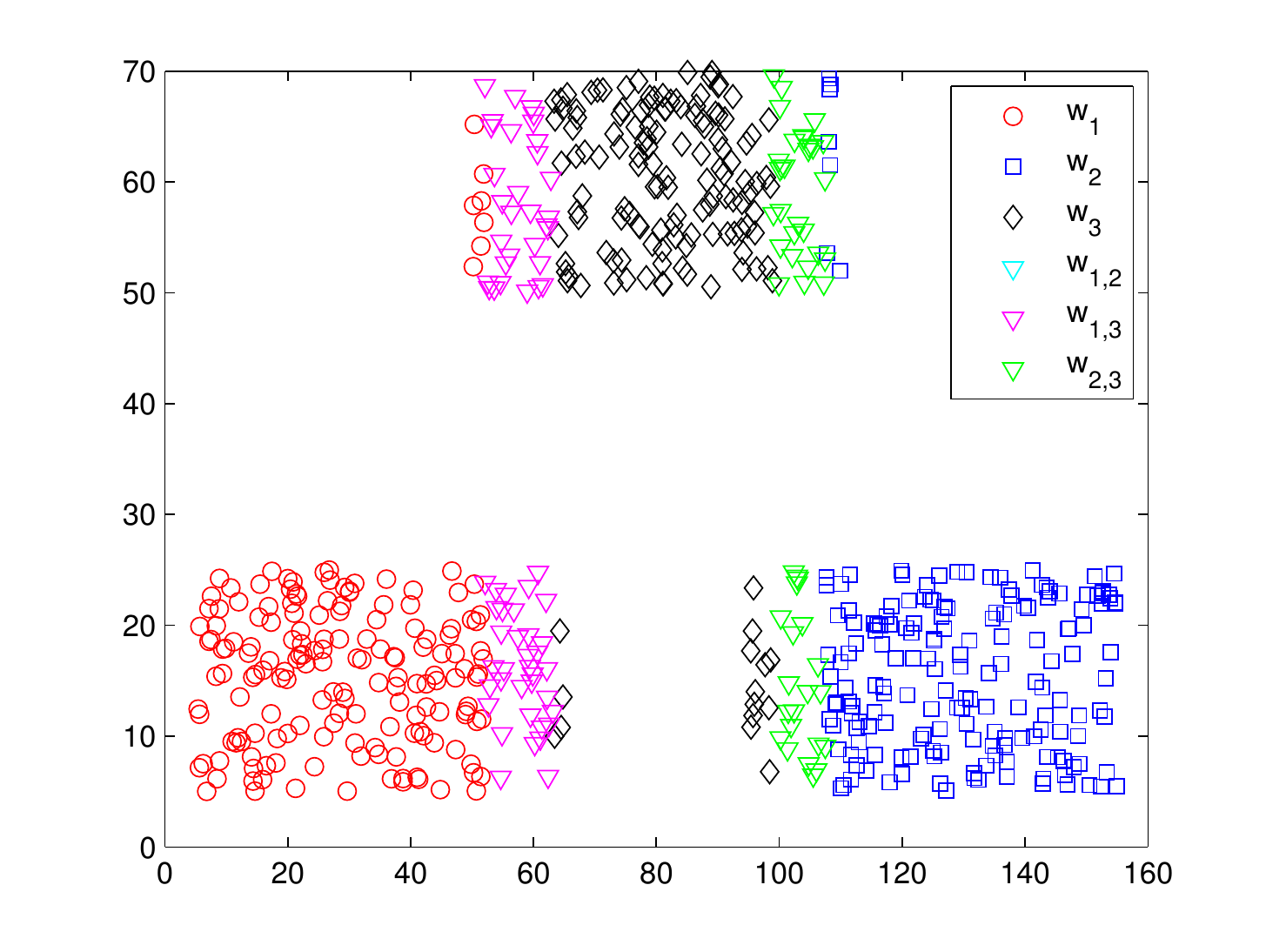} 
	\hfill
\includegraphics[width=.450\linewidth]{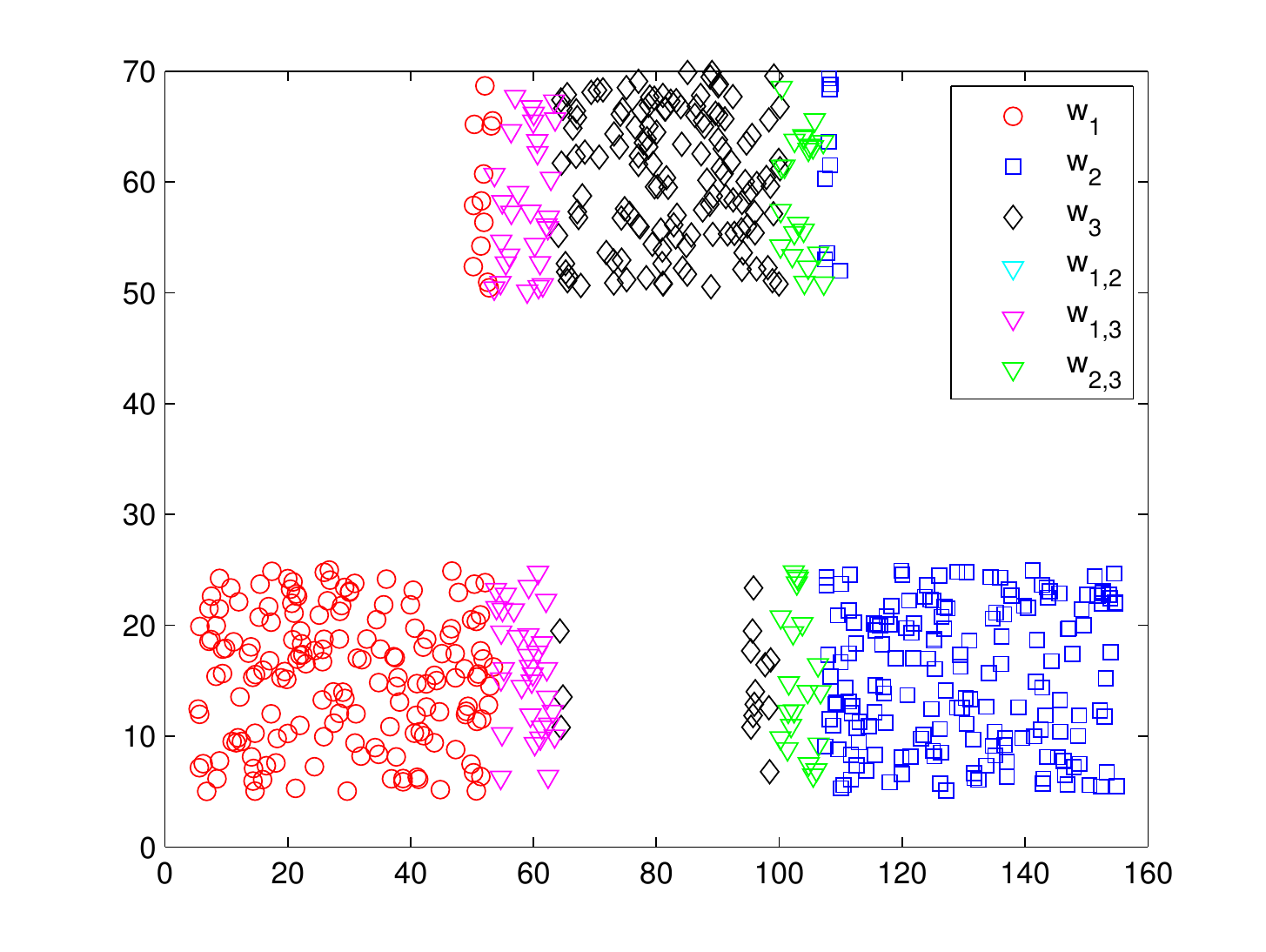}
\hfill
\parbox{.450\linewidth}{\centering\small (e). Classification result only by $2^{nd}$ step of SCCAI ($Re=4.83,Ri_2=19.33, time=0.1719s$). }
	\hfill
		\parbox{.450\linewidth}{\centering\small (f). Classification result by CCAI ($Re=5.83,Ri_2=16.83, time=0.0469s$).  }
	\hfill			
\caption{ Classification results of a 3-class artificial data set by different methods. \label{figEX1}}
\end{figure}
\end{center}

  Several different methods like FCMI, KNNI, SOMI have been applied here for comparison with CCAI as shown by Fig. \ref{figEX1}-(a)--\ref{figEX1}-(f). Particularly, the classification result obtained using the (first or second) single step of CCAI (denoted by SCCAI) are also given as in Fig. \ref{figEX1}-(d)--\ref{figEX1}-(e).
 In the first step of CCAI, the direct classification is done without imputation of missing value, whereas the object is classified with imputation of missing values in all incomplete patterns by the only second step of CCAI.

 A particular value of $K=9$ is selected in the classifier K-NN imputation method\footnote{In fact, the choice of $K$ ranking from 7 to 15 does not affect seriously the results.}. For notation conciseness, we have denoted $\omega^{te}\triangleq \omega^{test}$, $\omega^{tr}\triangleq \omega^{training}$ and $\omega_{i,\ldots, k}\triangleq \omega_i\cup \ldots\cup \omega_k$. The error rate (in \%), imprecision rate (in \%)  and computation time (Sec.) are specified in the caption of each subfigure.

Because the $y$ value in the test sample is missing, the class $w_3$ appears partially overlapped with the classes $\omega_1$ and $\omega_2$ on their margins according to the value of x-coordinate as shown in Fig. \ref{figEX1}-(a). The missing value of the samples in the overlapped parts can be filled by quite different estimations obtained from different classes with the almost same reliabilities. For example, the estimation of the missing values of the objects in the right margin of $\omega_1$ and the left margin of $\omega_3$ can be obtained according to the training class $\omega_1$ or $\omega_3$. The edited pattern with the estimation from $\omega_1$ will be classified into class $\omega_1$, whereas it will be committed to class $\omega_3$ if the estimation is drawn from $\omega_3$. It is similar to the test samples in the left margin of $\omega_2$  and the right margin of $\omega_3$. This indicates that the missing value play a crucial rule in the classification of these objects, but unfortunately the estimation of these involved missing values are quite uncertain according to context. So these objects are prudently classified into the proper meta-class (e.g. $\omega_1\cup \omega_3$ and $\omega_2\cup \omega_3$) by CCAI. The CCAI results indicate that these objects belong to one of the specific classes included in the meta-classes, but these specific classes cannot be clearly distinguished by the object based only on the available values. If one wants to get more precise and accurate classification results, one needs to request for additional resources for gathering more useful information. The other objects in the left margin of $\omega_1$, right margin of $\omega_2$ and middle of $\omega_3$ can be correctly classified based on the only known value in x-coordinate, and it is not necessary to estimate the missing value for the classification of these objects in CCAI. However, all the test samples are classified into specific classes by the traditional methods KNNI and FCMI, and this causes many errors due to the limitation of probability framework. If we just apply the first step of SCCAI without imputation of the missing value and directly classify all the objects using the only known value (i.e. value in x-coordinate), it produces bigger error rate than the other methods, and this indicates that the imputation procedure is important to improve the accuracy of classification. If only the second step of SCCAI is done with imputation of the missing values in all incomplete patterns, it causes high imprecision rate that is not an efficient solution, and it takes much longer computation time than CCAI. CCAI with the adaptive imputation strategy can well balance the error rate, imprecision rate and computation burden.
 CCAI consisting of two steps generally produces smaller error rate than KNNI, FCMI and SOMI thanks to the use of meta-classes. Meanwhile, the computational time of CCAI is similar to that of FCMI, and is much shorter than KNNI because of the introduction of SOM technique in the estimation of missing values. It shows that the computational complexity of CCAI is relatively low. This simple example shows the interest and the potential of the credal classification obtained with CCAI method.

%==============================
\subsection{Experiment 2 (artificial data set)}
%==============================

 In this second experiment, we evaluate the performance of CCAI method using a 4D data set which includes 3 classes  $\omega_1$, $\omega_2$, and $\omega_3$. The artificial data are generated from three 4D Gaussian distributions characterized by the following means vectors and covariance matrices ($\mathbf{I}$ denotes the $4\times 4$ identity matrix):
\begin{align*}
& \mathbf{\mu}_1=(10, 50, 100, 100)^T,  \mathbf{\Sigma}_1=10\cdot\mathbf{I}\\
& \mathbf{\mu}_2=(30, 40, 50, 90)^T,  \mathbf{\Sigma}_2=15\cdot\mathbf{I}\\
& \mathbf{\mu}_3=(20, 80, 90, 130)^T, \mathbf{\Sigma}_3=12\cdot\mathbf{I}
\end{align*}
We have used $g$ training samples, and $g$ test samples (for $g=500$, and $g=1000$) in each class. So there are total $N=3\times g$ training samples and $N=3\times g$ test samples. Each test sample has $n$ missing values (for $n=1,2,3$), and the missing component value is randomly distributed in every dimension.
Three other methods KNNI, FCMI, SOMI and PCC are also applied here for the performances comparison.
 For each pair $(N, n)$, the reported error rates, imprecision rates and running time (sec.) are the averages over 10 trials performed with 10 independent random generation of the data sets. For KNNI, the values of $K$ ranging from 5 to 20 neighbors have been tested, and the mean error rate with $K\in[5, 20]$ is given in Table \ref{Table2}. In PCC method, the parameter $\epsilon$ has been optimized to obtain an acceptable compromise between error rate and the imprecision degree. ENN is adopted to classify the edited pattern with imputation of missing values in FCMI, KNNI, SOMI and PCC.

\begin{center}
  \begin{table}[!h]
%  \footnotesize
 %   \scriptsize
    \small
\centering
\caption{Classification results for 3-class data set by different methods (in \%).}
\vspace{2mm}
\begin{tabular}{cccccc }
\hline
(N,n)& FCMI&KNNI & SOMI & PCC& CCAI\\
&$\{Re, time\}$&$\{Re, time\}$  &$\{Re, time\}$ &$\{Re, Ri_2, time\}$ & $\{Re, Ri_2, time\}$\\
  \hline
(1500,1)& \{6.73, 0.9094s\} &\{7.42, 3.0005s\} &\{7.22, 0.9814s\} &\{6.20, 2.33, 0.3484s\}& \{4.64, 3.87, 0.2500s\}\\
(1500,2)& \{14.38, 0.9016s\} &\{15.68, 2.7759s\} &\{15.43, 0.9546s\}& \{13.47, 5.93, 0.3141s\}& \{9.76, 9.79, 0.2344s\}\\
(1500,3)& \{36.84, 0.9391s\} &\{40.11, 3.002s\} &\{40.10, 1.0322s\}& \{34.57, 7.97, 0.3484s\}& \{29.71, 15.6, 0.2906s\}\\
\hline
(3000,1)& \{6.75, 1.3922s\} &\{7.54, 12.0386s\} &\{7.14, 1.7310s\}& \{6.17, 1.63, 0.5453s\}& \{4.73, 3.83, 0.3469s\}\\
(3000,2)& \{14.73, 1.5375s\} &\{15.80, 11.3857s\} &\{15.20, 1.8203s\}& \{14.00, 1.60, 0.5234s\}& \{9.90, 10.33, 0.3063s\}\\
(3000,3)& \{36.43,1.6500s\} &\{40.48, 10.2803s\} &\{40.05, 1.6094s\}&\{33.94, 8.13, 0.5484s\}& \{29.52, 16.83, 0.3937s\}\\
 \hline
 \end{tabular}
\label{Table2}
\end{table}
\end{center}

The classification results of the applied methods (i.e. FCMI, KNNI, SOMI, PCC and CCAI) have been shown in Table \ref{Table2}. Our proposed CCAI method produces the lowest error rate, since some objects hard to correctly classify because of the missing values have been committed to the proper meta-class. Meanwhile, CCAI takes the shortest computation time compared with the other methods. This is because that some incomplete patterns are directly classified ignoring the missing values, which are considered unimportant for the classification. However, the missing values in each pattern are all imputed by other methods, and this needs more computations and thus increases the computational time. Moreover, one can see that KNNI takes the longest time, and this is the main drawback of K-NN based method. The K-NN strategy is also adopted in CCAI, but we use a few optimized  weighting vectors acquired by SOM technique to represent the whole training data class. Thus, we just need to calculate the distances between the object and these obtained  weighting vectors rather than all the training samples, which reduces a lot the computation burden.

%==============================
\subsection{ Experiment 3 (real data set)}
%==============================
Nine well known real data sets\footnote{We select seven classes from Yeast data set, because the last three classes (i.e. VAC POX and ERL) contain quite few samples.} available from UCI Machine Learning Repository \cite{UCI} are used in this experiment to evaluate the performance of CCAI with respect to KNNI, FCMI, SOMI and PCC. Both ENN and EK-NN are employed here as standard classifier to classify the edited patterns. Moreover, the single ($1^{st}$ and $2^{nd}$) step procedure of CCAI (SCCAI) has been also applied here for comparison. In the first step of SCCAI, the object is directly classified using the only available attributes without imputation procedure, whereas all the missing values are imputed before the classification in the second step of SCCAI. The basic information of these used real data sets is given in Table \ref{Table_data}. In Hepatitis data set, many patterns have already contained missing values. The patterns with missing values are considered as test samples, and the others are used as training samples.
There is no missing values in the other seven original data sets, and it is assumed that $n$ values are missing completely at random in all dimensions of each test sample. The cross validation is performed for these seven data sets, and we use the simplest 2-fold cross validation\footnote{More precisely, the samples in each class are randomly assigned to two sets $S_1$ and $S_2$ having equal size. Then we train on $S_1$ and test on $S_2$, and reciprocally. } here, since it has the advantage that the training and test sets are both large, and each sample is used for both training and testing on each fold.
 The 2-fold cross validation has been repeated 10 times, and the average error rate $Re$ and imprecision rate $Ri$ (for PCC and CCAI) of the different methods are given in Table \ref{Table_UCI_results}. Particularly, the reported classification result of KNNI is the average with  $K$ value ranging from 5 to 15.
 For the notation conciseness, the selected classifier (SC) is denoted by A=EK-NN, B=ENN in Table \ref{Table_UCI_results}. For the method of single step of CCAI (SCCAI), A represents the first step of SCCAI, and B represents the second step of SCCAI.

  \begin{table}[!h]
\centering
\caption{ Basic information of the used data sets. }
\vspace{0.5mm}
\begin{tabular}{cccc}
\hline
 name & classes & attributes & instances\\
 \hline
Breast & 2 &9&699\\
Hepatitis& 2 & 19 &155\\
Statlog (Heart)&2 &13& 270\\
Iris& 3&4&150\\
Seeds& 3 &7&210\\
Wine& 3 & 13& 178 \\
Knowledge & 4&5&403\\
Vehicle&4& 18&946\\
Yeast&7 &8&1429\\
\hline
\end{tabular}
\label{Table_data}
\end{table}

\begin{center}
  \begin{table}[!h]
\centering
\caption{ Classification results for different real data sets (rates in \%). }
\vspace{1mm}
\begin{tabular}{lccccccc}
\hline
data set& (n,SC)&FCMI& KNNI &SOMI &PCC& SCCAI& CCAI\\
& & $Re$& $Re$  & $Re$  &$\{Re, Ri_2\}$  &$\{Re, Ri_2\}$&$\{Re, Ri_2\}$\\
\hline
 Hepatitis &A&26.40&27.38&27.47&\{22.22, 7.56\}&\{23.67, 0\}&\{21.33, 5.33\}\\
&B&25.33&26.67&25.33&\{20.00, 6.67\}&\{20.00, 8.00\}&\\
\hline
&(3,A)&3.96&4.83&3.85&\{4.39, 2.20\}&\{4.98, 0\}&\{3.66, 0\}\\
&(3,B)&3.81& 3.95&3.51&\{3.81, 2.34\} &\{3.22, 0.73\}& \\
&(6,A)& 6.18&9.07&6.47&\{5.82, 1.93\}&\{6.15, 0\}&\{4.83, 1.61\} \\
Breast&(6,B)&7.32& 8.20&5.93&\{5.42, 1.32\}&\{4.72, 2.93\}& \\
&(7,A)& 12.02&14.00&13.62&\{10.11, 2.86\}&\{12.15, 0\}&\{9.00, 0.66\}\\
&(7,B) &11.42 &  11.54&12.45&\{10.10, 2.64\}&\{7.03, 17.11\}&\\
\hline
&(1,A)&6.89&5.29&5.14&\{4.80, 2.04\}&\{6.67, 0\}&\{4.00, 1.33\} \\
&(1,B)&7.33& 4.89&5.00&\{5.33, 2.67\} &\{4.00, 3.33\}&\\
&(2,A)&13.89&13.02&13.24&\{8.31, 6.27\}&\{12.00, 0\}&\{8.00, 4.67\} \\
Iris&(2,B)&14.00& 11.33&12.67&\{8.67, 4.00\} &\{7.33, 8.00\}&\\
&(3,A)& 18.22& 18.67&18.00&\{13.33, 8.67\}&\{17.33, 0\}&\{11.33, 12.00\}\\
&(3,B)&17.33 & 18.44&17.34&\{12.67, 9.33\}&\{10.67, 16.00\}&\\
\hline
&(2,A)&15.56&11.59&11.63&\{10.51, 2.95\}&\{9.52, 0\}&\{9.52, 0\}\\
&(2,B)&15.24& 11.19&  10.20  &\{9.52, 4.76 \} &\{9.52, 0.95\}& \\
&(4,A)&18.17&12.70&12.86&\{10.22, 3.52\}&\{10.48, 0\}&\{10.00, 0.48\}\\
Seeds&(4,B)&17.14& 11.98&12.59&\{10.48, 4.29\} &\{9.52, 1.90\}& \\
&(6,A)& 21.75&26.41&25.65&\{17.84, 10.32\}&\{22.86, 0\}&\{16.19, 13.81\}\\
&(6,B) &20.95 &  25.71&24.63&\{16.19, 14.76\}&\{8.10, 28.57\}&\\
\hline
&(3,A)& 29.32&27.12&27.53&\{27.38, 0.71\}&\{6.97, 0\}&\{6.74, 1.12\}\\
&(3,B)&26.97& 26.97&28.65&\{26.97, 1.69\} &\{6.18, 8.43\}& \\
&(7,A)&34.68&26.22&31.30&\{27.12, 0.79\}&\{7.87, 0\}&\{7.30, 3.93\}\\
Wine&(7,B)&33.24&30.43&31.46&\{29.78, 2.25\} &\{5.62, 9.55\}& \\
&(11,A)& 34.76 &29.55 &34.35&\{29.06, 1.61\}&\{14.61, 0\}&\{12.36, 3.93\}\\
&(11,B) &33.43 &  30.90&32.58&\{30.34, 2.81\}&\{10.67, 40.45\}&\\
\hline
&(1,A)&30.07&28.53&29.78&\{26.72, 4.05\} &\{27.55, 0\}&\{ 20.85, 6.20\} \\
&(1,B)&34.50&33.51&33.88&\{28.35, 6.31\} &\{20.10, 8.19\}&\\
 Knowledge &(2,A)&33.06&29.66&31.51&\{27.32, 5.36\}& \{30.69, 0\}&\{23.57, 6.95\}\\
&(2,B)&39.68&39.43&41.69&\{33.32, 7.73\}&\{20.35, 13.40\}\\
&(3,A)&34.32&32.96&35.24&\{29.86, 9.97\}&\{34.16, 0\}&\{30.51, 7.69\}\\
&(3,B)&39.96&40.69&42.04&\{33.76, 11.82\}&\{22.08, 21.59\}&\\
\hline
&(1,A)& 37.41&37.78&36.67&\{33.41, 12.59\}&\{17.78, 0\}&\{16.30, 0.37\}\\
 Heart &(1,B)& 41.18&41.85&41.11&\{36.30, 9.63\}&\{13.70, 21.48\}&\\
&(5,A)&48.15&38.27&41.48&\{35.06, 25.93\}&\{23.70, 0\}& \{22.96, 0.74\}\\
&(5,B)&46.89&43.09&42.96&\{32.96, 28.52\}&\{22.59, 8.89\}\\
\hline
 &(5,A)&46.00&41.13&41.25&\{35.63, 25.75\}&\{50.71, 0\}&\{34.87, 26.48\}\\
 Vehicle &(5,B)&56.66&55.67&54.73&\{37.87, 27.43\}&\{27.66, 50.24\}&\\
&(9,A)&57.97&45.27&45.68&\{38.63, 22.73\}&\{52.25, 0\}&\{36.64, 22.34\}\\
&(9,B)&61.82&57.92&57.71&\{43.63, 26.95\}&\{28.61, 56.97\}&\\
\hline
&(1,A)&46.57&46.04&45.51&\{42.71, 11.12\}&\{46.67, 0\}&\{40.28, 12.36\}\\
 Yeast &(1,B)&44.97&44.72&44.86&\{39.86, 13.92\}&\{27.08, 46.74\}\\
&(3,A)&54.29&54.22&54.88&\{51.86, 10.87\}&\{56.74, 0\}&\{49.75, 12.64\}\\
&(3,B)&51.72&52.81&53.89&\{49.38, 13.69\}&\{34.38, 49.31\}&\\
\hline
 \end{tabular}
\label{Table_UCI_results}
\end{table}
\end{center}

\vspace{5mm}
One can see in  Table \ref{Table_UCI_results} that the credal classification of PCC and CCAI always produce the lower error rate than the traditional FCMI, KNNI and SOMI methods, since some objects that cannot be correctly classified using only the available attribute values have been properly committed to the meta-classes, which can well reveal the imprecision of classification. The selected classifiers (i.e. EK-NN and ENN) for classification of edited patterns in FCMI, KNNI, SOMI and PCC are usually with the similar performance in many cases\footnote{EK-NN outperforms ENN sometimes, but ENN can be better in some other cases.}, but it is known that the K-NN based method generally has big computation burden. The choice of EK-NN and ENN should be made according to the actual condition in real applications.
In CCAI, some objects with the imputation of missing values are still classified into the meta-class. It indicates that these missing values play a crucial role in the classification, but the estimation of these missing values is no very good. In other words, the missing values can be filled with the similar reliabilities by different estimated data, which lead to distinct classification results. So we have to cautiously assign them to the meta-class to reduce the risk of misclassification. Compared with our previous method PCC, this new method CCAI generally provide better performance with lower error rate and imprecision rate, and it is mainly because more accurate estimation method (i.e. SOM+KNN) for missing values is adopted in CCAI. However, if only the first step of SCCAI is applied, it produces more misclassification errors that other methods  due to the absence of imputation of missing data. Whereas, the imprecision rate will be quite high if only the second step of SCCAI is adopted because all the conflicting beliefs caused in the combination procedure are transferred to the meta-classes. So CCAI with adaptive imputation of missing values can provide a good compromise between the error and imprecision. This third experiment using real data sets shows the effectiveness and interest of this new CCAI method with respect to other methods.

%===============================================
\section{Conclusion}
%===============================================

A new credal classification method with adaptive imputation of missing values (called CCAI) for dealing with incomplete pattern has been presented based on belief function theory. In the first step of CCAI method, some objects (incomplete pattern) are directly classified ignoring the missing values if the specific classification result can be obtained, which effectively reduces the computation complexity because it avoids the imputation of the missing values. However, if the available information is not sufficient to achieve a specific classification of the object in the first step, we estimate (recover) the missing values before entering the classification procedure in a second step. The SOM and K-NN techniques are applied to make the estimation of missing attributes with a good compromise between the estimation accuracy and computation burden. The credal classification in this work allows the object to belong to different singleton classes and meta-class (i.e. disjunction of several classes) with different masses of belief. Once the object is committed to a meta-class (e.g. $A\cup B$), it means that the missing values cannot be accurately recovered according to the context, and the estimation is not very good. Different estimations will lead the object to distinct classes (e.g. $A$ or $B$) involved in the meta-class. So some other sources of information will be required to achieve more precise classification of the object if necessary. The credal classification is able to well capture the imprecision of classification thanks to the meta-class and it effectively reduces the misclassification errors. The effectiveness and interest of the proposed CCAI method have been evaluated on three distinct experiments using artificial and real data sets.

\bibliographystyle{model3-num-names}

\end{document}